\documentclass[11pt]{article}

\usepackage[preprint]{acl}

\usepackage{times}
\usepackage{latexsym}

\usepackage[T1]{fontenc}

\usepackage[utf8]{inputenc}

\usepackage{microtype}

\usepackage{inconsolata}

\usepackage{graphicx}

\usepackage{enumitem}
\usepackage{booktabs}
\usepackage{amsmath}
\usepackage{amssymb}
\usepackage{mathtools}
\usepackage{amsthm}
\usepackage{makecell}
\usepackage{pifont}

\theoremstyle{plain}
\newtheorem{theorem}{Theorem}[section]

\theoremstyle{definition}

\theoremstyle{remark}

\NewDocumentCommand{\var}{O{s} m O{}}{%
  \ensuremath{#1_{#2}^{#3}}
}
\usepackage{siunitx}
\usepackage{multirow}
\usepackage[table]{xcolor}
\usepackage{array}
\definecolor{highlightred}{RGB}{244, 204, 204}
\definecolor{highlightgreen}{RGB}{204, 244, 204}
\definecolor{highlightyellow}{RGB}{255, 255, 204}

\definecolor{dropgreen}{HTML}{C8E6C9}    
\definecolor{dropyellow}{HTML}{FFF9C4}   
\definecolor{droporange}{HTML}{FFE0B2}   
\definecolor{dropred}{HTML}{FFCDD2}      

\newcommand{\dropcolor}[1]{%
  \pgfmathtruncatemacro{\dropband}{%
    abs(#1) < 1 ? 1 : (abs(#1) < 2 ? 2 : (abs(#1) < 4 ? 3 : 4))%
  }%
  \ifnum\dropband=1
    \cellcolor{dropgreen!80!white}\ensuremath{#1}%
  \else
    \ifnum\dropband=2
      \cellcolor{dropyellow!80!white}\ensuremath{#1}%
    \else
      \ifnum\dropband=3
        \cellcolor{droporange!80!white}\ensuremath{#1}%
      \else
        \cellcolor{dropred!80!white}\ensuremath{#1}%
      \fi
    \fi
  \fi
}

\newcommand{\defeq}{\ensuremath{\mathrel{\mathop:}=}}

\newcommand{\commentout}[1]{}

\definecolor{light-gray}{gray}{0.80}

\newcommand\fref{Fig.~\ref}
\newcommand\tref{Table~\ref}

\usepackage{amsthm}

\usepackage{amsthm}

\newtheorem{mytheorem}{Theorem}
\newtheorem{assumption}[mytheorem]{Assumption}
\newtheorem{lemma}[mytheorem]{Lemma}
\newtheorem{remk}[mytheorem]{Remark}

\usepackage{xspace}

\usepackage[capitalize,noabbrev]{cleveref}

\usepackage[most]{tcolorbox}

\newcommand{\mean}[1]{\ensuremath{#1}}

\definecolor{rqgreen}{RGB}{0,120,0}
\definecolor{rqbrown}{RGB}{120,70,20}
\definecolor{deeporange}{RGB}{210,105,30}
\newtcolorbox{rqblock}{
  enhanced,
  colback=deeporange!12,
  colframe=deeporange,
  boxrule=1.2pt,
  arc=8pt, 
  left=5pt,right=5pt,top=5pt,bottom=5pt,
}

\title{Quantization Inflates Reasoning: Token Inflation as a Hidden Cost of Low-Bit Reasoning Models}

\author{
  \textbf{Xinyu Lian\textsuperscript{1}},
  \textbf{Walid Krichene\textsuperscript{2}},
  \textbf{Beichen Huang\textsuperscript{1}},
  \textbf{Masahiro Tanaka\textsuperscript{3}},
\\
  \textbf{Olatunji Ruwase\textsuperscript{4}},
  \textbf{Li Zhang\textsuperscript{2}},
  \textbf{Minjia Zhang\textsuperscript{1}}
\\
  \textsuperscript{1}University of Illinois Urbana-Champaign,
  \textsuperscript{2}Microsoft,
  \textsuperscript{3}Anyscale,
  \textsuperscript{4}Snowflake
}

\begin{document}

\maketitle

\begin{abstract}


Quantization is widely used to reduce the inference cost of large language models, but its effect on reasoning models is not fully captured by final-answer accuracy or per-token latency. We show that low-bit post-training quantization can introduce a hidden test-time compute cost: quantized reasoning models often generate longer chains of thought even when they still answer correctly. 
Across mathematical reasoning, code generation, scientific question answering, and agentic tool-use benchmarks, we find that INT4/INT3 quantization can preserve accuracy but increase reasoning-token usage, offsetting the expected per-token speedup. To measure this effect, we introduce the CoT Token Inflation Ratio, which compares reasoning length between quantized and full-precision models averaged across all evaluation benchmarks. We further show that token inflation is accompanied by behavioral changes in the reasoning trace, including more intermediate steps and greater semantic repetition. These changes translate into measurable end-to-end real-world serving penalties. Finally, we evaluate mitigation strategies and find that prompting and decoding-time sampling offer inconsistent accuracy-length trade-offs, while quantization-aware training shows more promise in reducing both accuracy degradation and token inflation. Our results suggest that reasoning-token usage should be reported alongside accuracy when evaluating quantized reasoning models. 


\end{abstract}




\section{Introduction}
\label{sec:intro}


Recent large language models (LLMs) trained for step-by-step reasoning have achieved strong performance on mathematical reasoning, code generation, and scientific question answering~\citep{deepseek_r1,o1_system_card}.
A key driver of this progress is the ability to spend more test-time computation: instead of producing a short answer directly, reasoning models generate long chains of thought (CoT) that decompose the problem, explore intermediate steps, verify partial results, and sometimes revise earlier conclusions~\citep{lightman2023verify,deepseek_r1,o1_system_card,kimi_k15_2025}.
This behavior improves accuracy on difficult tasks, but it also makes inference substantially more expensive. For reasoning models, deployment cost is no longer determined only by model size and per-token latency. It also depends on how many reasoning tokens the model chooses to generate. 


Quantization is one of the most widely used techniques for reducing the inference cost of LLMs.  Post-training quantization (PTQ) methods such as GPTQ~\cite{gptq2022}, AWQ~\cite{awq2023}, SmoothQuant~\cite{smoothquant}, ParoQuant~\cite{paroquant} reduce model memory footprint and accelerate decoding with negligible training, and they have shown strong accuracy-efficiency trade-offs on LLM benchmarks. These results have made low-bit quantization a practical default for serving large models. However, most quantization evaluations measure final-answer accuracy and per-token efficiency. This evaluation protocol is incomplete for reasoning models, where the number of generated reasoning tokens is a major component of test-time compute. 


In this paper, we show that low-bit post-training quantization can introduce a hidden cost for reasoning models: it can make models reason longer. Across mathematical reasoning, code generation, scientific question answering, and agentic tool-use benchmarks, we find that quantized models often preserve final-answer accuracy while generating longer chains of thought than their full-precision counterparts. This effect can offset, and in some cases outweigh, the per-token speedup expected from quantization. In other words, a quantized reasoning model can be faster per decoding step but slower end-to-end because it takes more tokens to reach the answer. 


We call this phenomenon \textbf{reasoning-token inflation} and introduce the CoT Token Inflation Ratio (CTIR) to measure it. CTIR measures the relative change in reasoning-token count between a quantized model and its full-precision reference, averaged across all evaluation benchmarks. 


Our analysis further shows that token inflation is not merely harmless verbosity. Quantized models often produce more intermediate reasoning steps and more semantically repetitive traces, revisiting or re-verifying prior reasoning without improving correctness. These behavior changes translate into measurable end-to-end serving cost penalties. We also examine preliminary mitigation strategies, finding that prompting and decoding-time controls provide inconsistent accuracy-length trade-offs, while quantization-aware training (QAT) is a more promising direction for reducing both accuracy degradation and reasoning-token inflation. 
In summary, this paper makes three contributions:



\begin{itemize}[leftmargin=*]
    \item We study PTQ on SoTA reasoning models ranging from 4B to 30B parameters (dense and MoE) and quantitatively identify reasoning-token inflation as a hidden cost of low-bit post-training quantization. Across reasoning, code, and agentic tool-use benchmarks, quantized models can preserve final-answer accuracy while generating longer chains of thought, offsetting the expected per-token efficiency gains (\S\ref{sec:hidden_cost}).    

    \item We introduce the CoT Token Inflation Ratio (CTIR), a metric for measuring quantization-induced increases in reasoning length relative to a full-precision reference. CTIR exposes an evaluation gap in standard quantization studies, which typically report accuracy and per-token latency but omit the amount of reasoning generated at inference time  (\S\ref{sec:inflation_rootcause}). 



    \item We analyze the behavioral consequences of token inflation. Quantized models often produce more intermediate reasoning steps and more semantically repetitive traces, and this increased reasoning length translates into higher end-to-end serving latency and cost (\S\ref{sec:why_inflation}). We further evaluate mitigation strategies, finding that prompting and decoding-time sampling offer inconsistent accuracy-length trade-offs, while quantization-aware training is a promising direction (\S\ref{sec:mitigation}).
\end{itemize}

\section{Background and Related Work}
\label{sec:background}

\subsection{Quantization Methods for LLM Deployment}

\noindent
\textbf{Quantization for LLM deployment.} Post-training quantization (PTQ) and quantization-aware training (QAT) are two main paradigms for reducing LLM inference cost. PTQ converts a pretrained model to low precision with little or no additional training~\citep{hqq,gptq2022,awq2023,autoround,paroquant}, while QAT simulates low-precision arithmetic during training or fine-tuning with quantize--dequantize operators~\citep{esser2020lsq,liu2024llmqat,chee2025discquant}. Because training or fine-tuning large LLMs is expensive, most prior work has focused on PTQ methods that avoid additional training cost~\citep{optrot,quant-error-prop,ptq-microscaling,spinquant,quipsharp,omniquant2023}. Representative methods include ZeroQuant~\citep{zeroquant}, GPTQ~\citep{gptq2022}, AWQ~\citep{awq2023}, SmoothQuant~\citep{smoothquant}, QuIP\#~\citep{quipsharp}, and SpinQuant~\citep{spinquant}. These methods reduce memory footprint and memory traffic through reconstruction-based quantization, activation-aware scaling, outlier smoothing, rotations, or low-bit codebooks, and have become attractive for LLM deployment due to their accuracy--efficiency trade-offs. QAT can better adapt weights to low-bit constraints, but its training cost remains substantial at LLM scale, making PTQ the more common practical choice.

\noindent
\textbf{Quantization for reasoning models.}
Recent work has begun to study how quantization affects reasoning-heavy tasks. \citet{li2025quantization_meets_reasoning} evaluate common PTQ methods, including GPTQ, AWQ, and SmoothQuant, on mathematical reasoning and show that extreme low-bit settings can substantially degrade accuracy. They further propose lightweight fine-tuning for recovery. \citet{liu2025quantization_reasoning} broaden the study across model families and quantization targets, including weights, activations, and KV cache, and observe that harder reasoning tasks suffer larger quantization-induced degradation. These studies show that reasoning models are more sensitive to quantization than standard LLM benchmarks reveal.
Our work studies a different, deployment-critical effect. Prior work primarily evaluates final-answer accuracy, with limited analysis of how quantization changes reasoning length. \citet{liu2025quantization_reasoning} consider CoT length in a limited setting of distilled models. In contrast, we analyze quantized reasoning models across a range of scales and architectures, including both dense and MoE models, and jointly study accuracy, reasoning-token usage, trace-level behavior, and serving efficiency. We show that PTQ can preserve final-answer accuracy while inflating the reasoning CoT span, creating a hidden test-time compute cost that accuracy-only evaluation misses. We further evaluate mitigation strategies spanning inference-time controls, reasoning-aligned calibration data, and quantization-aware training, showing that reducing token inflation requires preserving the model's reasoning trajectory rather than merely shortening its output.


\section{Experimental Setup}
\label{sec:task_setup}

We evaluate how low-bit post-training quantization (PTQ) affects both the accuracy and generation behavior of reasoning models.

\noindent
\textbf{Models.}
We evaluate open-source reasoning models spanning dense and mixture-of-experts (MoE) architectures. The dense models include Qwen3-4B-Thinking-2507 (Qwen3-4B), Phi-4-Reasoning-Plus (Phi-4-14B). The MoE models include Qwen3-30B-A3B-Thinking (Qwen3-30B).
This model set covers a broad range of scales, from small reasoning models to large MoE reasoning models, allowing us to study whether quantization-induced changes in reasoning behavior manifest across model size and architecture. The full model checkpoint information and architectural details are provided in Appendix~\ref{app:model}.

\noindent
\textbf{Benchmarks.}
Our benchmark suite covers mathematical reasoning, code generation, scientific reasoning, broad multi-step reasoning, and general language understanding. 
For mathematical reasoning, we use AIME25 and GSM8K. These benchmarks span grade-school word problems and high-difficulty competition problems. For code generation, we use LiveCodeBench (LCB) and MBPP, evaluating generated programs by functional correctness against held-out tests. For scientific and broad-knowledge reasoning, we use GPQA-Diamond.
We additionally include BBH as a mixed multi-step reasoning benchmark. To compare against less reasoning-intensive workloads, we report HellaSwag, MMLU-Pro and PIQA as general benchmarks.
Each benchmark is evaluated on 100 questions sampled via random shuffling (seed 42); AIME25 is evaluated on its full set of 30 questions. Benchmark details are provided in Appendix~\ref{app:benchmarks}.
All outputs are generated using \texttt{vLLM} with temperature 0 to remove sampling variance and isolate the effect of quantization on the model's generation trajectory.

\noindent
\textbf{Quantization Methods.}
We focus on weight-only PTQ, a common low-bit deployment setting for LLM inference. This scope lets us isolate a widely used quantization setting and study whether lower-precision weights alone can change the reasoning behavior of LLMs. 
We evaluate representative PTQ methods, including \textit{RTN} (round-to-nearest, the zero-cost baseline with no calibration), \textit{GPTQ} (second-order, reconstruction-based weight quantization)~\citep{gptq2022}, \textit{AWQ} (activation-aware weight quantization)~\citep{awq2023}, \textit{HQQ} (calibration-free weight quantization)~\citep{hqq},
and \textit{ParoQuant} (pairwise Givens-rotation quantization with channel-wise scaling that flattens per-channel magnitudes)~\citep{paroquant}.
For each method, we apply INT4 and INT3 quantization with group size 128. INT4 represents a practical low-bit deployment target, while INT3 provides a more aggressive setting for testing whether reasoning-token inflation becomes stronger as quantization noise increases. BF16 inference serves as the full-precision reference throughout the paper.


\section{When Accuracy Preservation Fails to Predict Efficiency}
\label{sec:hidden_cost}

\newcommand{\modelrot}[2][1.95cm]{%
  \rotatebox[origin=c]{90}{\parbox[c]{#1}{\centering #2}}%
}

\newcommand{\rothdr}[1]{%
  \rotatebox[origin=c]{90}{\bfseries #1}%
}

\newcommand{\na}{\text{---}} 
\begin{table*}[t]
    \centering
    \fontsize{4.5pt}{5.5pt}\selectfont

    
    \setlength{\tabcolsep}{6.5pt}
    \renewcommand{\arraystretch}{1.2}

    \begin{tabular}{l|l|l|c c c c c c|c|c|c c c|c|c}
    \toprule
    \multirow{2}{*}{\rothdr{Model}} & \multirow{2}{*}{\textbf{Prec.}} & \multirow{2}{*}{\textbf{Quant.}} &
    \multicolumn{8}{c|}{\textbf{Reasoning Tasks}} &
    \multicolumn{5}{c}{\textbf{General Tasks}} \\
    \cmidrule(lr){4-11}\cmidrule(lr){12-16}
    & & &
    \textbf{AIME25} & \textbf{GSM8K} & \textbf{GPQA-D} & \textbf{BBH} & \textbf{LCB} & \textbf{MBPP} &
    \textbf{Avg-R} & \textbf{Drop-R$\downarrow$} &
    \textbf{HellaSwag} & \textbf{PIQA} & \textbf{MMLU-Pro} &
    \textbf{Avg-G} & \textbf{Drop-G$\downarrow$} \\
    \midrule
    \midrule

   \multirow{10}{*}{\modelrot{Qwen3-4B}}
    & BF16 & --- &
    \mean{76.7} & \mean{94.0} & \mean{70.0} & \mean{87.0} & \mean{15.0} & \mean{92.0} & \mean{72.5} & --- & \mean{50.0} & \mean{81.0} & \mean{60.0} & \mean{63.7} & --- \\
    \cline{2-16}
    & \multirow{5}{*}{INT4} & RTN  &
    \mean{50.0} & \mean{91.0} & \mean{58.0} & \mean{84.0} & \mean{19.0} & \mean{65.0} & \mean{61.2} & \dropcolor{11.3} & \mean{51.0} & \mean{83.0} & \mean{62.0} & \mean{65.3} & \dropcolor{-1.7} \\
    & & GPTQ &
    \mean{73.3} & \mean{95.0} & \mean{68.0} & \mean{87.0} & \mean{16.0} & \mean{90.0} & \mean{71.6} & \dropcolor{0.9} & \mean{49.0} & \mean{77.0} & \mean{56.0} & \mean{60.7} & \dropcolor{3.0} \\
    & & AWQ & 
    \mean{80.0} & \mean{95.0} & \mean{63.0} & \mean{89.0} & \mean{18.0} & \mean{94.0} & \mean{73.2} & \dropcolor{-0.7} & \mean{50.0} & \mean{79.0} & \mean{61.0} & \mean{63.3} & \dropcolor{0.3} \\
    & & HQQ &
    \mean{70.0} & \mean{95.0} & \mean{64.0} & \mean{89.0} & \mean{17.0} & \mean{89.0} & \mean{70.7} & \dropcolor{1.8} & \mean{50.0} & \mean{79.0} & \mean{58.0} & \mean{62.3} & \dropcolor{1.3} \\
    & & ParoQuant &
    \mean{80.0} & \mean{96.0} & \mean{64.0} & \mean{87.0} & \mean{19.0} & \mean{91.0} & \mean{72.8} & \dropcolor{-0.4} & \mean{50.0} & \mean{81.0} & \mean{61.0} & \mean{64.0} & \dropcolor{-0.3} \\

    \cline{2-16}
    & \multirow{4}{*}{INT3} & GPTQ & 
    \mean{63.3} & \mean{92.0} & \mean{52.0} & \mean{85.0} & \mean{22.0} & \mean{88.0} & \mean{67.1} & \dropcolor{5.4} & \mean{49.0} & \mean{76.0} & \mean{58.0} & \mean{61.0} & \dropcolor{2.7} \\
    & & AWQ & 
    \mean{20.0} & \mean{89.0} & \mean{44.0} & \mean{78.0} & \mean{2.0} & \mean{41.0} & \mean{45.7} & \dropcolor{26.8} & \mean{48.0} & \mean{76.0} & \mean{57.0} & \mean{60.3} & \dropcolor{3.3} \\
    & & HQQ  &
    \mean{26.7} & \mean{93.0} & \mean{54.0} & \mean{84.0} & \mean{9.0} & \mean{66.0} & \mean{55.4} & \dropcolor{17.0} & \mean{49.0} & \mean{78.0} & \mean{61.0} & \mean{62.7} & \dropcolor{1.0} \\
    & & ParoQuant &
    \mean{43.3} & \mean{96.0} & \mean{50.0} & \mean{84.0} & \mean{34.0} & \mean{91.0} & \mean{66.4} & \dropcolor{6.1} & \mean{50.0} & \mean{77.0} & \mean{57.0} & \mean{61.3} & \dropcolor{2.3} \\
    \midrule
    \midrule

    \multirow{8}{*}[1.8ex]{\modelrot{Phi4-plus-14B}}
    & BF16 & --- & 
    \mean{73.3} & \mean{44.0} & \mean{55.0} & \mean{17.0} & \mean{0.0} & \mean{39.0} & \mean{38.1} & --- & \mean{64.0} & \mean{82.0} & \mean{77.0} & \mean{74.3} & --- \\
    \cline{2-16}
    & \multirow{4}{*}{INT4} & RTN & 
    \mean{60.0} & \mean{26.0} & \mean{61.0} & \mean{16.0} & \mean{0.0} & \mean{38.0} & \mean{33.5} & \dropcolor{4.6} & \mean{62.0} & \mean{81.0} & \mean{76.0} & \mean{73.0} & \dropcolor{1.3} \\
    & & GPTQ & 
    \mean{53.3} & \mean{41.0} & \mean{64.0} & \mean{18.0} & \mean{0.0} & \mean{41.0} & \mean{36.2} & \dropcolor{1.9} & \mean{62.0} & \mean{81.0} & \mean{78.0} & \mean{73.7} & \dropcolor{0.6} \\
    & & AWQ & 
    \mean{53.3} & \mean{30.0} & \mean{69.0} & \mean{21.0} & \mean{0.0} & \mean{33.0} & \mean{34.4} & \dropcolor{3.7} & \mean{64.0} & \mean{82.0} & \mean{77.0} & \mean{74.3} & \dropcolor{0.0} \\
    & & HQQ & 
    \mean{60.0} & \mean{48.0} & \mean{66.0} & \mean{15.0} & \mean{0.0} & \mean{43.0} & \mean{38.7} & \dropcolor{-0.6} & \mean{65.0} & \mean{81.0} & \mean{78.0} & \mean{74.7} & \dropcolor{-0.3} \\
    \cline{2-16}
    & \multirow{3}{*}{INT3} & GPTQ & 
    \mean{50.0} & \mean{32.0} & \mean{51.0} & \mean{15.0} & \mean{0.0} & \mean{22.0} & \mean{28.3} & \dropcolor{9.7} & \mean{60.0} & \mean{82.0} & \mean{75.0} & \mean{72.3} & \dropcolor{2.0} \\
    & & AWQ & 
    \mean{50.0} & \mean{53.0} & \mean{59.0} & \mean{12.0} & \mean{0.0} & \mean{45.0} & \mean{36.5} & \dropcolor{1.6} & \mean{62.0} & \mean{81.0} & \mean{77.0} & \mean{73.3} & \dropcolor{1.0} \\
    & & HQQ &
    \mean{33.3} & \mean{37.0} & \mean{49.0} & \mean{12.0} & \mean{1.0} & \mean{36.0} & \mean{28.1} & \dropcolor{10.0} & \mean{61.0} & \mean{82.0} & \mean{74.0} & \mean{72.3} & \dropcolor{2.0} \\
    \midrule
    \midrule

    \multirow{8}{*}[1.5ex]{\modelrot{Qwen3-30B}}
    & BF16 & --- &
    \mean{86.7} & \mean{97.0} & \mean{71.0} & \mean{84.0} & \mean{21.0} & \mean{89.0} & \mean{74.8} & --- & \mean{59.0} & \mean{80.0} & \mean{80.0} & \mean{73.0} & --- \\
    \cline{2-16}
    & \multirow{4}{*}{INT4} & RTN &
    \mean{83.3} & \mean{95.0} & \mean{70.0} & \mean{90.0} & \mean{22.0} & \mean{93.0} & \mean{75.6} & \dropcolor{-0.8} & \mean{60.0} & \mean{82.0} & \mean{75.0} & \mean{72.3} & \dropcolor{0.7} \\
    & & GPTQ &
    \mean{86.7} & \mean{95.0} & \mean{74.0} & \mean{85.0} & \mean{17.0} & \mean{90.0} & \mean{74.6} & \dropcolor{0.2} & \mean{59.0} & \mean{80.0} & \mean{80.0} & \mean{73.0} & \dropcolor{0.0} \\
    & & AWQ &
    \mean{83.3} & \mean{96.0} & \mean{72.0} & \mean{90.0} & \mean{18.0} & \mean{85.0} & \mean{74.1} & \dropcolor{0.7} & \mean{59.0} & \mean{78.0} & \mean{78.0} & \mean{71.7} & \dropcolor{1.3} \\
    & & HQQ &
    \mean{86.7} & \mean{97.0} & \mean{74.0} & \mean{92.0} & \mean{27.0} & \mean{87.0} & \mean{77.3} & \dropcolor{-2.5} & \mean{60.0} & \mean{79.0} & \mean{81.0} & \mean{73.3} & \dropcolor{-0.3} \\
    \cline{2-16}
    & \multirow{3}{*}{INT3} & GPTQ &
    \mean{83.3} & \mean{94.0} & \mean{74.0} & \mean{90.0} & \mean{23.0} & \mean{84.0} & \mean{74.7} & \dropcolor{0.1} & \mean{56.0} & \mean{80.0} & \mean{78.0} & \mean{71.3} & \dropcolor{1.7} \\
    & & AWQ &
    \mean{73.3} & \mean{96.0} & \mean{64.0} & \mean{88.0} & \mean{15.0} & \mean{74.0} & \mean{68.4} & \dropcolor{6.4} & \mean{56.0} & \mean{78.0} & \mean{72.0} & \mean{68.7} & \dropcolor{4.3} \\
    & & HQQ &
    \mean{63.3} & \mean{95.0} & \mean{66.0} & \mean{85.0} & \mean{12.0} & \mean{65.0} & \mean{64.4} & \dropcolor{10.4} & \mean{56.0} & \mean{78.0} & \mean{78.0} & \mean{70.7} & \dropcolor{2.3} \\
    \bottomrule
    \end{tabular}
    \caption{Overall evaluation of 3-bit and 4-bit quantization for Qwen3 and Phi-4 on reasoning and general benchmarks.
    We report separate averages and drops for reasoning (Avg-R/Drop-R) and general (Avg-G/Drop-G) tasks.
    Green, yellow, orange, and red cells denote negligible ($<1$), small ($1$--$2$), moderate ($2$--$4$), and large ($>4$) accuracy drops, respectively. LCB denotes LiveCodeBench. *Phi4 scores 0\% on LCB due to harness/format mismatches.}
    \label{tab:quant-results-sheet}
    \vspace{-10pt}
\end{table*}


\noindent
\textbf{Low-bit PTQ often preserves final-answer accuracy.} We begin with a broad evaluation of PTQ on reasoning models. 
Table~\ref{tab:quant-results-sheet} summarizes accuracy across reasoning and general benchmarks. To make the trends easier to interpret, we categorize absolute accuracy drops relative to BF16 into four bands: \emph{negligible} (drop $<1$), \emph{small} ($1$--$2$), \emph{moderate} ($2$--$4$), and \emph{large} ($>4$).
At INT4, the majority of (model, quantizer, benchmark) cells fall into the negligible or small degradation bands. This is especially true for general benchmarks (e.g., HellaSwag, PIQA). Larger reasoning models are also relatively robust: for Qwen3-30B, most INT4 reasoning-task drops remain within the negligible-to-moderate range ($<4$ points). These results are consistent with the common view that INT4 PTQ can be a practical deployment strategy when judged by final-answer accuracy alone. 

\noindent
\textbf{Reasoning-heavy tasks are more sensitive to quantization.}
Despite the overall robustness of INT4, Table~\ref{tab:quant-results-sheet} also shows a clear gap between reasoning and general benchmarks. Under the same quantization setting, general-task accuracy usually changes little, while reasoning benchmarks exhibit larger drops and more frequent moderate or large degradations. The degradation is concentrated on harder, long-horizon tasks such as AIME25, where small perturbations can compound over many reasoning steps. This suggests that final-answer accuracy on general benchmarks is not sufficient to characterize quantization quality for reasoning models. 

\noindent
\textbf{Accuracy degradation grows under more aggressive quantization.}
The accuracy gap becomes more pronounced as precision decreases from INT4 to INT3. Smaller reasoning models are particularly sensitive: for example, Qwen3-4B suffers a large absolute drop on AIME25 under INT4, and INT3 further increases degradation across difficult reasoning tasks. Larger models are more tolerant of INT4 noise, but they are not immune to aggressive quantization. Across models, INT3 consistently widens the gap between reasoning and general benchmarks, indicating that reasoning-heavy workloads are a more sensitive stress test for low-bit quantization than general-purpose evaluations.

\noindent
\textbf{Accuracy preservation does not imply end-to-end efficiency.}
The above results might suggest that INT4 PTQ is broadly effective: accuracy is often preserved, especially for larger models, and low-bit weights should reduce memory traffic and improve per-token decoding efficiency. However, this conclusion is incomplete for reasoning models. \emph{End-to-end latency depends not only on the cost of generating each token, but also on the number of tokens generated before the final answer}. In our measurements, quantized reasoning models often fail to translate optimized kernel-level speedups into end-to-end gains, and in some cases even increase total generation time. For example, running BBH with Qwen3-4B takes 285s in BF16 but 292s with INT4 quantization. This is despite the commonly used W4A16 Marlin kernel achieving a 1.2--1.4$\times$ tokens/s speedup over BF16 at batch sizes 1--16. \footnote{Tested in \texttt{vLLM} with input\_length=512 and max\_new\_tokens=128.} 

This motivates the central question of the next section: does quantization change how much reasoning the model generates?

\section{Quantization Inflates Reasoning Tokens}
\label{sec:inflation_rootcause}

The previous section showed that low-bit PTQ can preserve final-answer accuracy but fail to deliver the expected end-to-end efficiency gains on reasoning workloads. We now show that the missing factor is output length. Quantized reasoning models often generate longer chains of thought than their full-precision counterparts. 

\providecommand{\modelrot}[2][1.95cm]{%
  \rotatebox[origin=c]{90}{\parbox[c]{#1}{\centering #2}}%
}
\providecommand{\rothdr}[1]{%
  \rotatebox[origin=c]{90}{\bfseries #1}%
}
\providecommand{\na}{\text{---}}

\begin{table*}[t]
    \centering
    \tiny
    
    \setlength{\tabcolsep}{4.5pt}
    \renewcommand{\arraystretch}{1.3}

    \resizebox{\textwidth}{!}{%
    \begin{tabular}{l|l|l|cc cc cc cc cc cc|c}
    \toprule
    \midrule
    \multirow{2}{*}{\rothdr{Model}} & \multirow{2}{*}{\textbf{Prec.}} & \multirow{2}{*}{\textbf{Quant.}} &
    \multicolumn{12}{c|}{\textbf{Reasoning Tasks}} &
    \multirow{2}{*}{\textbf{CTIR Avg.}} \\
    \cmidrule(lr){4-15}
    & & &
    \multicolumn{2}{c}{\textbf{AIME25}} &
    \multicolumn{2}{c}{\textbf{GSM8K}} &
    \multicolumn{2}{c}{\textbf{GPQA-D}} &
    \multicolumn{2}{c}{\textbf{BBH}} &
    \multicolumn{2}{c}{\textbf{LCB}} &
    \multicolumn{2}{c|}{\textbf{MBPP}} &
    \\
    \midrule
    \midrule

    \multirow{10}{*}{\modelrot{Qwen3-4B}}
    & BF16 & --- &
    \mean{21307} & \na & \mean{1139} & \na & \mean{7796} & \na & \mean{2674} & \na & \mean{13459} & \na & \mean{4692} & \na & \na \\
    \cline{2-16}
    & \multirow{5}{*}{INT4} & RTN &
    \mean{24178} & \mean{+13.5\%} & \mean{1548} & \mean{+35.9\%} & \mean{10596} & \mean{+35.9\%} & \mean{5470} & \mean{+104.6\%} & \mean{17574} & \mean{+30.6\%} & \mean{6310} & \mean{+34.5\%} & \mean{+42.5\%} \\
    & & AWQ &
    \mean{22263} & \mean{+4.5\%} & \mean{1108} & \mean{-2.7\%} & \mean{7248} & \mean{-7.0\%} & \mean{2866} & \mean{+7.2\%} & \mean{14930} & \mean{+10.9\%} & \mean{6400} & \mean{+36.4\%} & \mean{+8.2\%} \\
    & & GPTQ &
    \mean{24388} & \mean{+14.5\%} & \mean{1157} & \mean{+1.6\%} & \mean{8565} & \mean{+9.9\%} & \mean{3417} & \mean{+27.8\%} & \mean{13930} & \mean{+3.5\%} & \mean{5380} & \mean{+14.7\%} & \mean{+12.0\%} \\
    & & HQQ &
    \mean{21795} & \mean{+2.3\%} & \mean{1202} & \mean{+5.5\%} & \mean{7626} & \mean{-2.2\%} & \mean{2995} & \mean{+12.0\%} & \mean{13303} & \mean{-1.2\%} & \mean{5597} & \mean{+19.3\%} & \mean{+6.0\%} \\
    & & ParoQuant &
    \mean{21652} & \mean{+1.6\%} & \mean{1136} & \mean{-0.3\%} & \mean{7744} & \mean{-0.7\%} & \mean{2905} & \mean{+8.6\%} & \mean{14416} & \mean{+7.1\%} & \mean{5240} & \mean{+11.7\%} & \mean{+4.7\%} \\
    \cline{2-16}
    & \multirow{3}{*}{INT3} & AWQ &
    \mean{30924} & \mean{+45.1\%} & \mean{1529} & \mean{+34.3\%} & \mean{13876} & \mean{+78.0\%} & \mean{7599} & \mean{+184.2\%} & \mean{61590} & \mean{+357.6\%} & \mean{49609} & \mean{+957.4\%} & \mean{+276.1\%} \\
    & & GPTQ &
    \mean{25402} & \mean{+19.2\%} & \mean{1456} & \mean{+27.8\%} & \mean{8912} & \mean{+14.3\%} & \mean{3946} & \mean{+47.6\%} & \mean{23773} & \mean{+76.6\%} & \mean{9704} & \mean{+106.8\%} & \mean{+48.7\%} \\
    & & HQQ &
    \mean{26038} & \mean{+22.2\%} & \mean{2508} & \mean{+120.2\%} & \mean{16599} & \mean{+112.9\%} & \mean{7392} & \mean{+176.4\%} & \mean{49690} & \mean{+269.2\%} & \mean{54018} & \mean{+1051.3\%} & \mean{+292.1\%} \\
    & & ParoQuant &
    \mean{23814} & \mean{+11.8\%} & \mean{1241} & \mean{+9.0\%} & \mean{10148} & \mean{+30.2\%} & \mean{2883} & \mean{+7.8\%} & \mean{16124} & \mean{+19.8\%} & \mean{7918} & \mean{+68.8\%} & \mean{+24.5\%} \\
    \midrule
    \midrule

    \multirow{8}{*}{\modelrot{Qwen3-30B}}
    & BF16 & --- &
    \mean{18587} & \na & \mean{865} & \na & \mean{6071} & \na & \mean{1692} & \na & \mean{15374} & \na & \mean{3330} & \na & \na \\
    \cline{2-16}
    & \multirow{4}{*}{INT4} & RTN &
    \mean{23613} & \mean{+27.0\%} & \mean{1098} & \mean{+27.0\%} & \mean{7188} & \mean{+18.4\%} & \mean{2177} & \mean{+28.7\%} & \mean{28088} & \mean{+82.7\%} & \mean{5809} & \mean{+74.4\%} & \mean{+43.0\%} \\
    & & AWQ &
    \mean{20103} & \mean{+8.2\%} & \mean{996} & \mean{+15.2\%} & \mean{6743} & \mean{+11.1\%} & \mean{2096} & \mean{+23.9\%} & \mean{17063} & \mean{+11.0\%} & \mean{3902} & \mean{+17.2\%} & \mean{+14.4\%} \\
    & & GPTQ &
    \mean{18277} & \mean{-1.7\%} & \mean{912} & \mean{+5.5\%} & \mean{6671} & \mean{+9.9\%} & \mean{1853} & \mean{+9.5\%} & \mean{15355} & \mean{-0.1\%} & \mean{3302} & \mean{-0.8\%} & \mean{+3.7\%} \\
    & & HQQ &
    \mean{19844} & \mean{+6.8\%} & \mean{895} & \mean{+3.6\%} & \mean{6323} & \mean{+4.1\%} & \mean{1678} & \mean{-0.8\%} & \mean{15751} & \mean{+2.5\%} & \mean{3569} & \mean{+7.2\%} & \mean{+3.9\%} \\
    \cline{2-16}
    & \multirow{3}{*}{INT3} & AWQ &
    \mean{21827} & \mean{+17.4\%} & \mean{982} & \mean{+13.6\%} & \mean{7352} & \mean{+21.1\%} & \mean{2194} & \mean{+29.7\%} & \mean{22131} & \mean{+43.9\%} & \mean{5700} & \mean{+71.2\%} & \mean{+32.8\%} \\
    & & GPTQ &
    \mean{18857} & \mean{+1.5\%} & \mean{964} & \mean{+11.5\%} & \mean{7233} & \mean{+19.1\%} & \mean{1932} & \mean{+14.2\%} & \mean{17583} & \mean{+14.4\%} & \mean{5114} & \mean{+53.6\%} & \mean{+19.0\%} \\
    & & HQQ &
    \mean{21183} & \mean{+14.0\%} & \mean{1036} & \mean{+19.9\%} & \mean{8380} & \mean{+38.0\%} & \mean{2388} & \mean{+41.1\%} & \mean{30812} & \mean{+100.4\%} & \mean{8235} & \mean{+147.3\%} & \mean{+60.1\%} \\
    \midrule
    \bottomrule
    \end{tabular}%
    
    }
    \caption{Average CoT tokens and CTIR across reasoning tasks. For each task, the two columns report CoT tokens and CTIR, respectively.}
    \label{tab:cot-token-ctir}
    \vspace{-10pt}
\end{table*}


\vspace{2pt}
\noindent
\textbf{CoT Token Inflation Ratio.}
To quantify this effect, we introduce the \emph{CoT Token Inflation Ratio (CTIR)}.
Let \(M_{\mathrm{org}}\) be the BF16 model and \(M_q\) be its quantized counterpart.
For an instance \(x\), we define the CoT length as
\begin{equation}
\ell_{\mathrm{cot}}(M,x)\defeq \big|\texttt{tok}(y_{\mathrm{cot}})\big|.
\end{equation}
For a dataset \(\mathcal{D}\), we define CTIR as the relative increase in average CoT length:
{\footnotesize
\begin{equation}
\bar{\ell}(M;\mathcal{D}) \defeq \frac{1}{|\mathcal{D}|}\sum_{x\in\mathcal{D}} \ell_{\mathrm{cot}}(M,x).
\end{equation}

\begin{equation}
\label{eq:ctir}
\mathrm{CTIR}(M_q;\mathcal{D}) \defeq
\frac{
\bar{\ell}(M_q;\mathcal{D})-\bar{\ell}(M_{\mathrm{org}};\mathcal{D})
}{
\bar{\ell}(M_{\mathrm{org}};\mathcal{D})
}
\times 100\%.
\end{equation}
}



\vspace{4pt}
\noindent
\textbf{Token inflation is widespread across quantization settings.}
Table~\ref{tab:cot-token-ctir} summarizes CTIR across models, reasoning benchmarks, quantization methods, and bit widths. The main pattern is that low-bit PTQ often increases reasoning-token usage, and the effect becomes stronger as precision decreases. INT4 usually leads to mild but measurable inflation, while INT3 produces substantially larger increases. The trend is especially dramatic for Qwen3-4B: under INT3, AWQ-INT3 reaches an average CTIR of 276.1\%, and HQQ-INT3 reaches 292.1\%. Larger models show greater resilience but are not immune: for Qwen3-30B under INT3, AWQ and HQQ reach 32.8\% and 60.1\%, respectively, still substantial, but far below the smaller model. This suggests that reasoning-token inflation becomes more severe as quantization noise increases. Quantization methods also differ in the severity of inflation. GPTQ generally produces smaller CTIR than AWQ and HQQ, suggesting that reconstruction-based quantization better preserves the full-precision generation trajectory. However, GPTQ does not eliminate the effect: INT3 GPTQ still increases CTIR relative to INT4 in most settings. These results show that final-answer accuracy alone hides the cost of quantization for reasoning models, because a quantized model can remain correct while spending more reasoning tokens to reach the answer.


\begin{table}[!t]
\centering
\scriptsize
\begin{tabular*}{\linewidth}{@{\extracolsep{\fill}}llccc}
\toprule
\textbf{Model} & \textbf{Quant.}
& \textbf{Acc.}
& \textbf{CoT Tok.}
& \textbf{Tool Calls} \\
\midrule
\multirow{3}{*}{Qwen3-4B}
& BF16      & 52.12\% & 6556 & 7.56 \\
& GPTQ-INT4 & 50.88\% & 7425 & 7.85 \\
& GPTQ-INT3 & 31.38\% & 7455 & 7.36 \\
\midrule
\multirow{3}{*}{Qwen3-30B}
& BF16      & 55.88\% & 5351 & 7.44 \\
& GPTQ-INT4 & 54.12\% & 5488 & 7.57 \\
& GPTQ-INT3 & 47.25\% & 6611 & 7.46 \\
\bottomrule
\end{tabular*}
\caption{BFCLv4-Multiturn evaluation on Qwen3-4B and Qwen3-30B models.}
\label{tab:bfcl-multiturn}
\vspace{-10pt}
\end{table}

Reasoning-token inflation can be even more costly in agentic workloads, where a model may perform multiple rounds of reasoning interleaved with tool calls. In such settings, latency is affected by both the number of tool invocations and the amount of reasoning generated around each invocation. We evaluate Qwen3-4B and Qwen3-30B on BFCLv4-Multiturn using GPTQ with a 4096-token reasoning budget per step. Table~\ref{tab:bfcl-multiturn} presents the results and highlights that quantization causes clear token inflation in agentic workloads, while leaving tool-use behavior nearly unchanged. For example, GPTQ-INT4 increases Qwen3-4B's CoT tokens by 13.3\% with only a small change in tool calls, from 7.56 to 7.85. For Qwen3-30B, GPTQ-INT3 increases CoT tokens by 23.5\%, while tool calls remain almost unchanged, from 7.44 to 7.46. This indicates that the E2E latency increase under quantization mainly comes from additional generated reasoning tokens rather than more frequent tool invocations.

\section{Anatomy of Reasoning-Token Inflation}
\label{sec:why_inflation}

\begin{figure}[ht]
  \centering
  \includegraphics[width=0.9\linewidth]{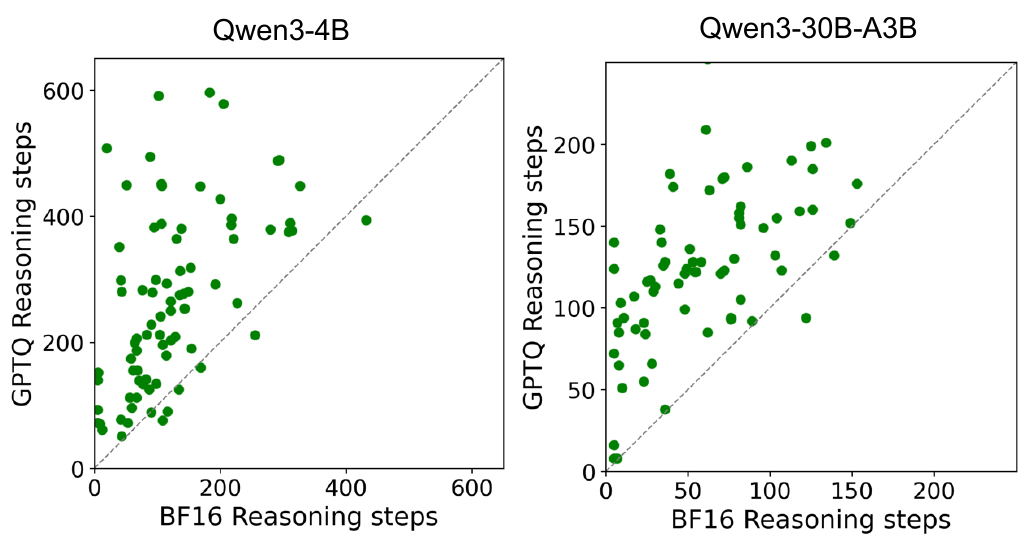}
  \caption{Per-instance CoT step counts under BF16 vs.\ INT3 GPTQ for Qwen3-4B (left) and Qwen3-30B (right) on MBPP. Each point plots BF16 steps (x-axis) against GPTQ steps (y-axis); the dashed line indicates parity. Points above the diagonal indicate more reasoning steps under quantization.}
  \label{fig:cot_step}
\end{figure}

\noindent
\textbf{Inflation appears as more reasoning steps.}
To unpack where the extra tokens come from, \fref{fig:cot_step} compares per-instance CoT step counts under BF16 and INT3 (GPTQ) for Qwen3-4B and Qwen3-30B on MBPP as an illustrative example. The majority of instances fall above the parity line, indicating that INT3 (GPTQ) generates more intermediate reasoning steps than BF16, suggesting that longer traces are not a byproduct of failures but reflect a systematic change in generation dynamics.

\noindent
\textbf{Inflation increases step-level repetition.}
Longer traces are not necessarily worse if the extra tokens correspond to useful additional reasoning. Manual inspection of reasoning traces reveals what appears to be an increase in \emph{repetitiveness} of the model. This can manifest in a number of ways: increased self-doubt and verification (e.g., ``wait'', ``hold on'', ``let me verify''), back-tracking and retries, or general repetition of prior CoT steps (e.g., ``again'', ``once more''). 
These behaviors resemble the reflective style of reasoning models, but under quantization they can appear more frequently without improving final-answer correctness.

To quantify this effect, we compute a step-level repetitiveness score. Given a CoT trace $c$ consisting of steps $c_1, \dots, c_k$, we measure a pairwise semantic similarity $s(c_i, c_j)$ by embedding each step using a sentence embedding model (specifically Qwen3-Embedding-4B), then computing the cosine similarity of each pair. We say that a step $c_i$ is repeated if it has at least one other step with a similarity exceeding a set threshold $\tau$.
Finally, the repetitiveness of $c$ is the fraction of repeated steps. Formally,
\[
\mathrm{Rep}(c_1, \dots, c_k) = \frac{1}{k} \big|\{i \mid \exists j \neq i : s(c_i, c_j) \geq \tau\}\big|.
\]

\begin{figure}[t]
  \centering
  \includegraphics[width=0.9\linewidth]{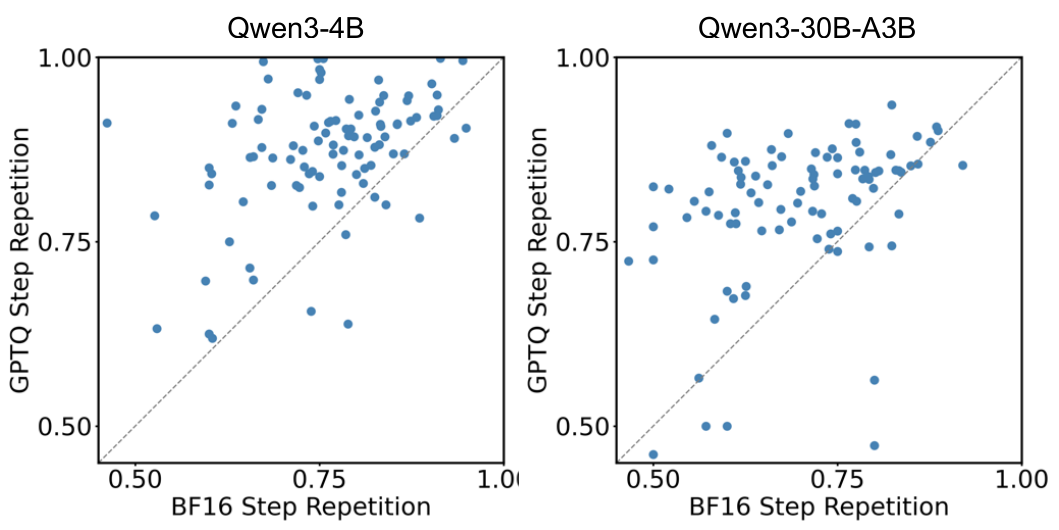}
  \caption{Per-instance CoT step repetitiveness (higher means more repetition) under BF16 vs.\ INT3 GPTQ for Qwen3-4B and Qwen3-30B on MBPP. Each point plots BF16 (x-axis) against INT3 GPTQ (y-axis); the dashed line indicates parity ($y=x$). Points above the diagonal indicate more repeated steps under quantization.}
  \label{fig:cot_step_repeat}
  \vspace{-10pt}
\end{figure}

\fref{fig:cot_step_repeat} compares per-instance repetitiveness under BF16 vs.\ INT3 (GPTQ) for Qwen3-4B and Qwen3-30B on MBPP, and shows that many instances lie above parity, indicating more near-duplicate steps under quantization. This increase in repetitiveness is consistent with the broader pattern that PTQ perturbs reasoning trajectories even when it does not immediately flip the final answer.

\vspace{2pt}
\noindent
\textbf{Modest token inflation can erase quantization speedups.}
The trace-level changes above have direct serving implications. In reasoning-model deployments, the chain of thought may be generated internally before the final answer is exposed to the user. In this setting, longer reasoning traces increase the time to first visible token, even if the per-token decoding cost is lower under quantization. Longer traces also occupy decoding slots for more steps and increase KV-cache residency time, reducing effective serving capacity under continuous batching.

\begin{figure}[t]
  \centering
  \includegraphics[width=0.95\linewidth]{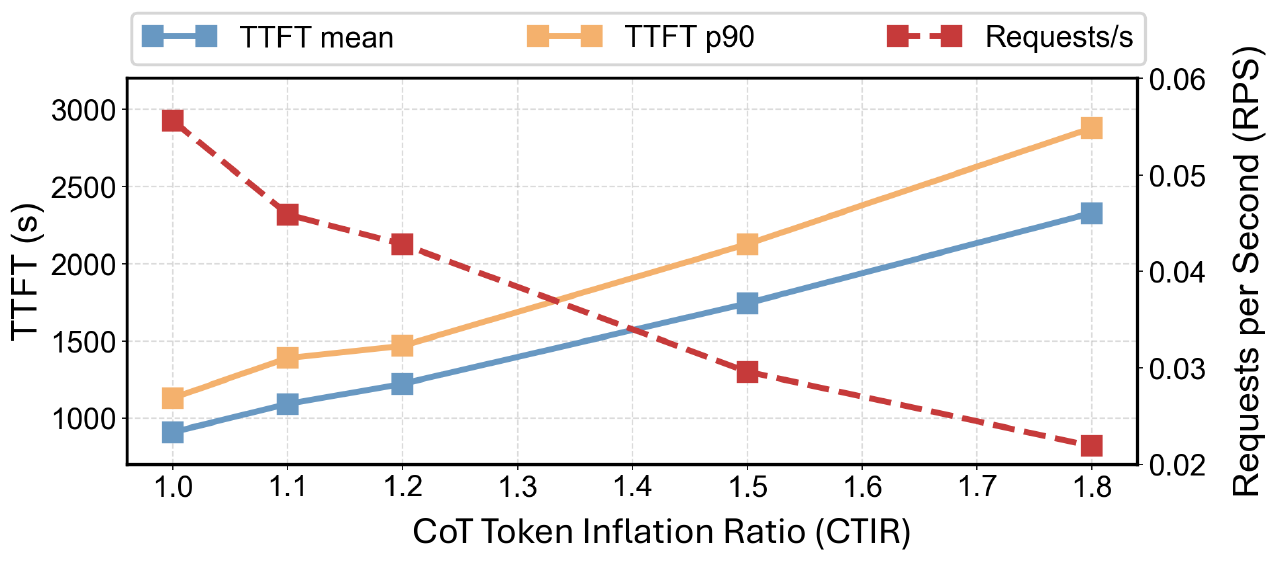}
  \caption{Impact of CoT length on serving efficiency. As reasoning length increases, time-to-first-token (TTFT; mean and p90) rises while \texttt{vLLM} throughput (requests/second) decreases.}
  \label{fig:ttft_throughput}
  \vspace{-10pt}
\end{figure}

We measure this effect using \texttt{vLLM}. We run Qwen3-30B-A3B on an H100 GPU with a 1k-token prompt and batch size 64. To isolate the effect of reasoning length, we control the CoT budget and measure time to first visible token (TTFT) and throughput. Here, the first visible token refers to the first user-facing output after the hidden reasoning span, so TTFT captures the user-perceived delay caused by generating the reasoning trace.
\fref{fig:ttft_throughput} shows that even modest CoT inflation can substantially affect serving performance.
A \(1.1\times\) CTIR increase raises p90 TTFT from 1050\,s to 1400\,s---an extra 350\,s (33\%) of latency for tail users. Longer generations also hold decoding slots and expand KV-cache footprints, squeezing concurrency under continuous batching: in the same setting, throughput drops from 0.055 to 0.047 requests/second. These results show why reasoning-token inflation matters in practice: the local efficiency gains of quantization can be offset when the quantized model generates more reasoning tokens.

\section{Can We Reduce Reasoning-Token Inflation?}
\label{sec:mitigation}

We now study whether token inflation can be reduced. We consider three classes of interventions with different costs: applying inference-time controls through decoding or prompting, choosing better calibration data for PTQ, and using quantization-aware training. 

\subsection{Inference-Time Controls Trade Accuracy for Shorter Reasoning}
A natural way to reduce reasoning token inflation is to intervene at inference time, either by discouraging repetition during decoding or by prompting the model to reason more concisely. These methods are attractive because they do not require retraining or re-quantizing the model. 

\begin{figure}[t]
  \centering  \includegraphics[width=0.95\linewidth]{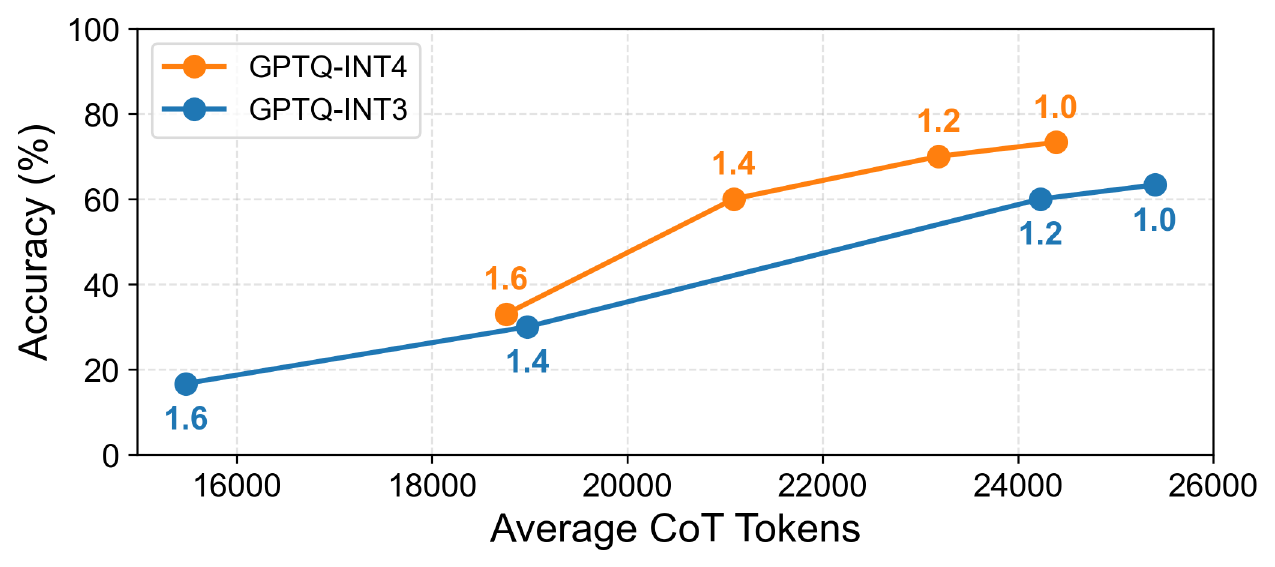}
  \vspace{-8pt}
  \caption{GPTQ quantized Qwen3-4B evaluation on AIME25, showing a trade-off between CoT token and accuracy under different repetition penalty settings.}
  \label{fig:tradeoff_aime}
  \vspace{-10pt}
\end{figure}

\paragraph{(i) Repetition penalty.}
We use the \texttt{repetition\_penalty} provided in \texttt{vLLM}, which discourages the model from generating tokens that have already appeared by rescaling their logits before sampling:
\[
z_i' =
\begin{cases}
z_i / r, & i \in \mathcal{S},\ z_i > 0, \\
z_i \cdot r, & i \in \mathcal{S},\ z_i \le 0, \\
z_i, & i \notin \mathcal{S},
\end{cases}
\]
where \(z_i\) is the original logit, \(r\) is the repetition penalty, and \(\mathcal{S}\) denotes the set of tokens that have appeared in the prompt or previously generated output. The default \texttt{repetition\_penalty} is 1, and we test values $r \in \{1.1, 1.2, 1.3, 1.4\}$ on Qwen3-4B and evaluate the performance on AIME25 as reported in Fig.~\ref{fig:tradeoff_aime}. Increasing the repetition penalty reduces the average CoT tokens, indicating that stronger repetition suppression can mitigate token inflation during reasoning. However, this token reduction often comes with a clear accuracy cost. These results suggest that repetition penalty is an effective but delicate control knob: it can reduce redundant reasoning tokens, yet overly strong penalties may also suppress useful intermediate reasoning steps and harm final-answer accuracy.

\paragraph{(ii) Conciseness prompting.}
Beyond decoding-time penalties, we further examine whether token inflation can be mitigated through prompting alone. Specifically, we test a set of lightweight prompt-based interventions that explicitly encourage shorter reasoning traces. The prompts are summarized as \textit{"do not think"}, \textit{"less tokens"}, and \textit{"no restate"}, and details are provided in Appendix~\ref{app:prompts}. We conduct this study on the same Qwen3-4B + GPTQ setup used in the sampling-penalty experiments and evaluate on MBPP. The results are reported in \tref{tab:prompt_intervention}, which shows that prompt-based interventions are a simple heuristic for reducing CoT length, but they often trade off accuracy and behave inconsistently. For example, \textit{no restate} cuts INT4 CoT tokens by 30.0\% but drops accuracy from 90.0\% to 81.0\%; for INT3, all prompts reduce tokens while causing 5–11 point accuracy drops. This suggests prompting is a heuristic, unstable mitigation for token inflation rather than a reliable solution.
\providecommand{\modelrot}[2][1.4cm]{%
  \rotatebox[origin=c]{90}{\parbox[c]{#1}{\centering #2}}%
}

\begin{table}[t]
\centering
\scriptsize
\setlength{\tabcolsep}{10pt}
\renewcommand{\arraystretch}{1.18}
\begin{tabular}{c c c c c}
\toprule
 & \textbf{Quant.} & \textbf{Prompt} & \textbf{Acc.} & \textbf{CoT Tok.} \\
\midrule
\multirow{8}{*}{\modelrot{MBPP}}
& \multirow{4}{*}{\modelrot[1.0cm]{INT4}}
& -- & 90.0\% & 5380 \\
& & \textit{do not think} & 90.0\% & 4496 \\
& & \textit{few tokens}   & 87.0\% & 5211 \\
& & \textit{no restate}   & 81.0\% & 3767 \\
\cmidrule(lr){2-5}
& \multirow{4}{*}{\modelrot[1.0cm]{INT3}}
& -- & 88.0\% & 9704 \\
& & \textit{do not think} & 83.0\% & 7494 \\
& & \textit{few tokens}   & 77.0\% & 6710 \\
& & \textit{no restate}   & 80.0\% & 7367 \\
\bottomrule
\end{tabular}
\caption{Prompt-based interventions on Qwen3-4B under GPTQ quantization for MBPP, reporting accuracy and CoT tokens. ``--'' denotes the baseline without additional prompt intervention.}
\label{tab:prompt_intervention}
\vspace{-15pt}
\end{table}



\subsection{Calibration Data as a Lightweight PTQ Mitigation}
Most PTQ methods rely on a small \emph{calibration} corpus to estimate activation statistics or solve layer-wise reconstruction objectives~\citep{gptq2022,awq2023,autoround}. While calibration data is known to affect accuracy, we show it simultaneously shapes reasoning-token inflation.

We compare five sources: (i) \texttt{ultrachat\_200k}, instruction dialogue~\citep{ultrachat200k}; (ii) \texttt{wikitext}~\citep{wikitext}; (iii) \texttt{c4}~\citep{T5}; (iv) \texttt{OpenR1-Math-220k}, math reasoning traces~\citep{openr1math220k}; and (v) \texttt{codeforces-cots}, competitive-programming reasoning traces; both (iv) and (v) are self-generated by the target BF16 model.

\fref{fig:compare} plots macro-average accuracy against average CoT length for Qwen3-4B under INT3 GPTQ.
Calibration data has a large joint effect on both axes: switching from Wikitext to Math-CoT raises accuracy from 28\% to 65\% while cutting CoT length from 30k to 12k tokens.
Reasoning-domain corpora (Math-CoT, Code-CoT) outperform generic text (Ultrachat, C4, Wikitext) on both dimensions.
Since this improvement requires no retraining beyond the standard PTQ procedure, we adopt Math-CoT as the default calibration corpus for all main-paper PTQ experiments.

\subsection{Quantization-Aware Training Achieves a Pareto Improvement}
\label{sec:qat}

\begin{figure}[t]
  \centering  \includegraphics[width=\linewidth]{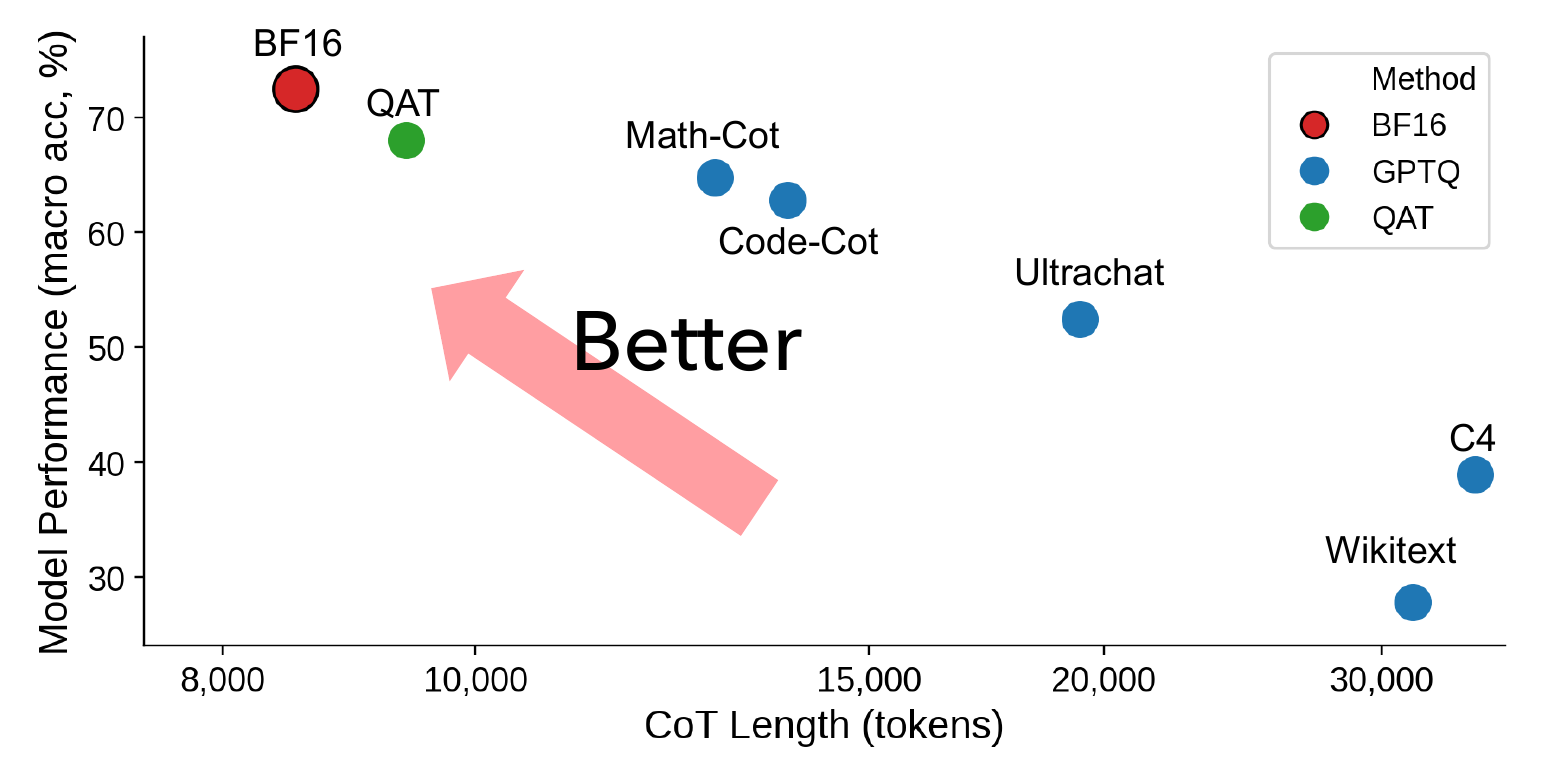}
  \vspace{-20pt}
  \caption{Accuracy--efficiency trade-off across calibration corpora for Qwen3-4B under INT3 GPTQ. Each point is a (GPTQ, corpus) pair; upper-left is better.}
  \label{fig:compare}
  \vspace{-10pt}
\end{figure}

We evaluate \textit{DiscQuant}~\citep{chee2025discquant}, a recent QAT method that jointly optimizes weight rounding under end-to-end task supervision,  achieving SoTA results.
In \fref{fig:compare}, QAT (green) reaches 68\% accuracy at 9.5k average CoT tokens---approaching BF16 (73\%, 8.5k tokens) and outperforming every PTQ configuration on both axes.

\vspace{2pt}
\noindent
\textbf{Why QAT can help.}
PTQ applies quantization post hoc to a model trained in full precision. Even when layer-wise reconstruction methods such as GPTQ reduce local quantization error, they do not directly optimize the model's behavior over long reasoning trajectories. Small perturbations can accumulate across many decoding steps, changing not only the final answer but also how long the model continues to reason. 
QAT provides a different signal. By back-propagating task loss through quantize-dequantize operators, QAT adapts the weights to low-bit execution rather than only correcting post-hoc rounding error. More importantly for reasoning models, when QAT is performed on reasoning traces, the model observes the target reasoning behavior under simulated quantization. This gives training signal not only for preserving final-answer correctness, but also for maintaining the structure and length scale of the full-precision reasoning trajectory. In this sense, QAT can re-align the quantized model with the reasoning process it is expected to preserve, whereas PTQ mainly attempts to minimize local reconstruction error. 

Among the three interventions, QAT is the only Pareto-dominant mitigation: it simultaneously reduces token inflation \emph{and} improves accuracy, while inference-time controls trade one for the other and calibration-data optimization remains strictly below QAT on both axes. The cost is a one-time retraining step.













\section{Conclusion}
\label{sec:conclusion}

Post-training quantization is widely used to reduce LLM inference cost, but our results show that accuracy alone is insufficient for evaluating quantized reasoning models. Low-bit PTQ can preserve final-answer accuracy while inflating chain-of-thought length, creating a hidden test-time compute cost that offsets expected per-token efficiency gains. We show that reasoning-token inflation is associated with longer and more repetitive reasoning traces, and that it increases time to first visible token and reduces serving throughput. These findings suggest that evaluations of quantized reasoning models should report reasoning-token usage alongside accuracy, especially when reasoning traces are hidden but still contribute to latency and serving capacity.





\section*{Acknowledgments}

This research was partially supported by the National Science Foundation (NSF) under Grant No. 2441601. The work utilized the Delta and DeltaAI system at the National Center for Supercomputing Applications (NCSA) and Jetstream2 at Indiana University through allocation CIS240055 from the Advanced Cyberinfrastructure Coordination Ecosystem: Services \& Support (ACCESS) program, which is supported by National Science Foundation grants \#2138259, \#2138286, \#2138307, \#2137603, and \#2138296. The Delta advanced computing resource is a collaborative effort between the University of Illinois Urbana-Champaign and NCSA, supported by the NSF (award OAC 2005572) and the State of Illinois. UIUC SSAIL Lab is supported by research funding and gift from Google, IBM, and AMD. 

\bibliography{reference,custom}

@article{paroquant,
  title={ParoQuant: Pairwise Rotation Quantization for Efficient Reasoning LLM Inference},
  author={Liang, Yesheng and Chen, Haisheng and Zhang, Zihan and Han, Song and Liu, Zhijian},
  journal={arXiv preprint arXiv:2511.10645},
  year={2025}
}

@inproceedings{autoround,
    title = "Optimize Weight Rounding via Signed Gradient Descent for the Quantization of {LLM}s",
    author = "Cheng, Wenhua and Zhang, Weiwei and Shen, Haihao and Cai, Yiyang and He, Xin and Kaokao, Lv and Liu, Yi",
    booktitle = "Findings of the Association for Computational Linguistics: EMNLP 2024",
    month = nov,
    year = "2024",
    address = "Miami, Florida, USA",
    publisher = "Association for Computational Linguistics",
    pages = "11332--11350",
    url = "https://aclanthology.org/2024.findings-emnlp.662/",
    doi = "10.18653/v1/2024.findings-emnlp.662",
    note = "Code: \url{https://github.com/intel/auto-round}"
}

@String{Computer = "{IEEE} Computer" }

@article{wikitext,
  author    = {Stephen Merity and
               Caiming Xiong and
               James Bradbury and
               Richard Socher},
  title     = {{Pointer Sentinel Mixture Models}},
  journal   = {CoRR},
  volume    = {arXiv preprint abs/1609.07843},
  year      = {2016}
}

@misc{openr1math220k,
  author       = {{open-r1 team}},
  title        = {{OpenR1-Math-220k}},
  howpublished = {\url{https://huggingface.co/datasets/open-r1/OpenR1-Math-220k}},
  year         = {2025},
  note         = {HuggingFace dataset. Accessed: 2026-01-23}
}

@misc{ultrachat200k,
  author       = {{HuggingFaceH4}},
  title        = {{UltraChat 200k}},
  howpublished = {\url{https://huggingface.co/datasets/HuggingFaceH4/ultrachat_200k}},
  year         = {2023},
  note         = {HuggingFace dataset. Accessed: 2026-01-23}
}

@article{T5,
  author    = {Colin Raffel and
               Noam Shazeer and
               Adam Roberts and
               Katherine Lee and
               Sharan Narang and
               Michael Matena and
               Yanqi Zhou and
               Wei Li and
               Peter J. Liu},
  title     = {Exploring the Limits of Transfer Learning with a Unified Text-to-Text
               Transformer},
  journal   = {CoRR},
  volume    = {abs/1910.10683},
  year      = {2019},
  url       = {http://arxiv.org/abs/1910.10683},
  archivePrefix = {arXiv},
  eprint    = {1910.10683},
  bibsource = {dblp computer science bibliography, https://dblp.org}
}

@inproceedings{cot,
  author       = {Jason Wei and
                  Xuezhi Wang and
                  Dale Schuurmans and
                  Maarten Bosma and
                  Brian Ichter and
                  Fei Xia and
                  Ed H. Chi and
                  Quoc V. Le and
                  Denny Zhou},
  Xeditor       = {Sanmi Koyejo and
                  S. Mohamed and
                  A. Agarwal and
                  Danielle Belgrave and
                  K. Cho and
                  A. Oh},
  title        = {Chain-of-Thought Prompting Elicits Reasoning in Large Language Models},
  booktitle    = {Advances in Neural Information Processing Systems 35: Annual Conference
                  on Neural Information Processing Systems 2022, NeurIPS 2022, New Orleans,
                  LA, USA, November 28 - December 9, 2022},
  year         = {2022}
}

@article{bbh,
  title="{Challenging BIG-Bench Tasks and Whether Chain-of-Thought Can Solve Them}",
  author={Suzgun, Mirac and Scales, Nathan and Sch{\"a}rli, Nathanael and Gehrmann, Sebastian and Tay, Yi and Chung, Hyung Won and Chowdhery, Aakanksha and Le, Quoc V. and Chi, Ed H. and Zhou, Denny and Wei, Jason},
  journal={arXiv preprint arXiv:2210.09261},
  year={2022}
}

@article{gsm8k,
  title="{Training Verifiers to Solve Math Word Problems}",
  author={Cobbe, Karl and Kosaraju, Vineet and Bavarian, Mohammad and Chen, Mark and Jun, Heewoo and Kaiser, {\L}ukasz and Plappert, Matthias and Tworek, Jerry and Hilton, Jacob and Nakano, Reiichiro and Hesse, Christopher and Schulman, John},
  journal={arXiv preprint arXiv:2110.14168},
  year={2021}
}

@article{gptq2022,
  title="{GPTQ: Accurate Post-Training Quantization for Generative Pre-trained Transformers}",
  author={Frantar, Elias and Ashkboos, Saleh and Hoefler, Torsten and Alistarh, Dan},
  journal={arXiv preprint arXiv:2210.17323},
  year={2022}
}

@article{awq2023,
  title="{AWQ: Activation-aware Weight Quantization for LLM Compression and Acceleration}",
  author={Lin, Ji and Tang, Jiaming and Tang, Haotian and Yang, Shang and Chen, Wei-Ming and Wang, Wei-Chen and Xiao, Guangxuan and Dang, Xingyu and Gan, Chuang and Han, Song},
  journal={arXiv preprint arXiv:2306.00978},
  year={2023}
}

@article{kimi_k15_2025,
  title="{Kimi k1.5: Scaling Reinforcement Learning with LLMs}",
  author={{Kimi Team}},
  journal={arXiv preprint arXiv:2501.12599},
  year={2025}
}

@article{omniquant2023,
  title="{OmniQuant: Omnidirectionally Calibrated Quantization for Large Language Models}",
  author={Shao, Wenqi and Chen, Mengzhao and Zhang, Zhaoyang and Xu, Peng and Zhao, Lirui and Li, Zhiqian and Zhang, Kaipeng and Gao, Peng and Qiao, Yu and Luo, Ping},
  journal={arXiv preprint arXiv:2308.13137},
  year={2023}
}

@inproceedings{esser2020lsq,
  title     = {Learned Step Size Quantization},
  author    = {Esser, Steven K. and McKinstry, Jeffrey L. and Bablani, Deepika and Appuswamy, Rathinakumar and Modha, Dharmendra S.},
  booktitle = {ICLR},
  year      = {2020}
}

@inproceedings{chee2025discquant,
  title     = "{DiscQuant: A Quantization Method for Neural Networks Inspired by Discrepancy Theory}",
  author    = {Chee, Jerry and Backurs, Arturs and Heck, Rainie and Zhang, Li and Kulkarni, Janardhan and Rothvoss, Thomas and Gopi, Sivakanth},
  booktitle = {Proceedings of Thirty Eighth Conference on Learning Theory (COLT 2025)},
  year      = {2025}
}

@article{liu2025quantization_reasoning,
  title="{Quantization Hurts Reasoning? An Empirical Study on Quantized Reasoning Models}",
  author={Liu, Ruikang and Sun, Yuxuan and Zhang, Manyi and Bai, Haoli and Yu, Xianzhi and Yu, Tiezheng and Yuan, Chun and Hou, Lu},
  journal={arXiv preprint arXiv:2504.04823},
  year={2025}
}

@article{li2025quantization_meets_reasoning,
  title="{Quantization Meets Reasoning: Exploring and Mitigating Degradation of Low-Bit LLMs in Mathematical Reasoning}",
  author={Li, Zhen and Su, Yupeng and Wang, Songmiao and Yang, Runming and Xie, Congkai and Liu, Aofan and Li, Ming and Cao, Jiannong and Xie, Yuan and Wong, Ngai and Yang, Hongxia},
  journal={arXiv preprint arXiv:2505.11574},
  year={2025}
}

@article{lightman2023verify,
  title={Let’s Verify Step by Step},
  author={Lightman, Hunter and Kosaraju, Vineet and Burda, Yura and Edwards, Harri and Baker, Bowen and Lee, Teddy and Leike, Jan and Schulman, John and Sutskever, Ilya and Cobbe, Karl},
  journal={arXiv preprint arXiv:2305.20050},
  year={2023}
}

@article{o1_system_card,
  title="{OpenAI o1 System Card}",
  author={OpenAI},
  journal={arXiv preprint arXiv:2412.16720},
  year={2024}
}

@article{deepseek_r1,
  title="{DeepSeek-R1: Incentivizing Reasoning Capability in LLMs via Reinforcement Learning}",
  author={{DeepSeek-AI}},
  journal={arXiv preprint arXiv:2501.12948},
  year={2025}
}

@inproceedings{liu2024llmqat,
  title     = "{{LLM-QAT}: Data-Free Quantization Aware Training for Large Language Models}",
  author    = {Liu, Zechun and Oguz, Barlas and Zhao, Changsheng and Chang, Ernie and Stock, Pierre and Mehdad, Yashar and Shi, Yangyang and Krishnamoorthi, Raghuraman and Chandra, Vikas},
  booktitle = {Findings of the Association for Computational Linguistics (ACL 2024)},
  year      = {2024}
}

@article{zeroquant,
  title="{ZeroQuant: Efficient and Affordable Post-Training Quantization for Large-Scale Transformers}",
  author={Yao, Zhewei and Yazdani Aminabadi, Reza and Zhang, Minjia and Wu, Xiaoxia and Li, Conglong and He, Yuxiong},
  journal={arXiv preprint arXiv:2206.01861},
  year={2022}
}

@misc{hqq,
  title        = "{Half-Quadratic Quantization of Large Machine Learning Models}",
  author       = {Badri, Hicham and Shaji, Appu},
  howpublished = {\url{https://dropbox.github.io/hqq_blog/}},
  note         = {Blog post},
  year         = {2023}
}

@article{smoothquant,
  title="{SmoothQuant: Accurate and Efficient Post-Training Quantization for Large Language Models}",
  author={Xiao, Guangxuan and Lin, Ji and Seznec, Micka{\"e}l and Wu, Hao and Demouth, Julien and Han, Song},
  journal={arXiv preprint arXiv:2211.10438},
  year={2022}
}

@article{quipsharp,
  title="{QuIP\#: Even Better LLM Quantization with Hadamard Incoherence and Lattice Codebooks}",
  author={Tseng, Albert and Chee, Jerry and Sun, Qingyao and Kuleshov, Volodymyr and De Sa, Christopher},
  journal={arXiv preprint arXiv:2402.04396},
  year={2024}
}

@article{spinquant,
  title="{SpinQuant: LLM Quantization with Learned Rotations}",
  author={Liu, Zechun and Zhao, Changsheng and Fedorov, Igor and Soran, Bilge and Choudhary, Dhruv and Krishnamoorthi, Raghuraman and Chandra, Vikas and Tian, Yuandong and Blankevoort, Tijmen},
  journal={arXiv preprint arXiv:2405.16406},
  year={2024}
}

@inproceedings{optrot,
  author       = {Advait Gadhikar and
                  Riccardo Grazzi and
                  James Hensman},
  title        = {OptRot: Mitigating Weight Outliers via Data-Free Rotations for Post-Training
                  Quantization},
  booktitle = {Advances in Neural Information Processing Systems},
  year      = {2025}
}

@inproceedings{quant-error-prop,
  author       = {Yamato Arai and
                  Yuma Ichikawa},
  title        = "{Quantization Error Propagation: Revisiting Layer-Wise Post-Training
                  Quantization}",
  booktitle = {Advances in Neural Information Processing Systems (NeurIPS 25)},
  year      = {2025}
}

@inproceedings{ptq-microscaling,
  author       = {Sayeh Sharify and
                  Utkarsh Saxena and
                  Zifei Xu and
                  Wanzin Yazar and
                  Ilya Soloveychik and
                  Xin Wang},
  title        = {Post Training Quantization of Large Language Models with Microscaling
                  Formats},
  booktitle = {Advances in Neural Information Processing Systems},
  year      = {2025}
}

\appendix
\section{Additional background on LLM quantization}
\label{app:quant_background}


\subsection{LLM quantization}
\label{subsec:back_quant}

\paragraph{Quantization preliminaries.}
Most practical LLM quantization methods use affine (uniform) quantization, mapping a floating-point tensor \(x\) to an integer tensor \(q\) with scale \(s>0\) and zero-point \(z\):
{\small
\begin{equation}
q = \mathrm{clip}\big(\mathrm{round}(x/s) + z,\ q_{\min}, q_{\max}\big), \
\hat{x} = s\cdot(q-z),
\end{equation}
}
where \(\hat{x}\) denotes the dequantized approximation of \(x\).
For Transformer weights, deployment-oriented pipelines typically adopt \emph{weight-only} quantization (e.g., INT4 weights with FP16/BF16 activations) because activation and KV-cache quantization can be more sensitive and often require kernel/hardware support. In addition, most LLM methods quantize weights \emph{per-channel} (or per-group) rather than per-tensor to reduce outlier-induced errors; in 4-bit regimes, group-wise quantization (e.g., group size 32/64/128) is common to balance accuracy and metadata overhead~\cite{gptq2022,awq2023,omniquant2023}.

\paragraph{Post-training quantization (PTQ).}
PTQ converts a pretrained model to low precision without full backpropagation-based training. It is attractive because it is efficient, requires limited calibration data (sometimes only a few hundred sequences), and can be applied to a frozen checkpoint.
Modern PTQ methods for LLMs largely fall into two families:

\emph{(i) Second-order / error-compensating weight quantization.}
GPTQ introduced a practical post-training procedure that quantizes weights layer-by-layer while approximately accounting for output error using curvature information (typically via a Hessian approximation from calibration data) \citep{gptq2022}. The key idea is to choose quantized weights that minimize an output reconstruction objective rather than performing naive round-to-nearest (RTN). GPTQ and its descendants (including highly optimized kernels in popular tooling) are widely used for INT4/INT3 weight-only deployment.

\emph{(ii) Saliency- and activation-aware weight quantization.}
AWQ observes that not all weights contribute equally to the output distribution; it protects ``important'' channels/weights by rescaling activations and applying clipping/quantization rules that reduce error on salient subspaces \citep{awq2023}. In practice, AWQ is often robust at INT4 for instruction-following models, and it is commonly paired with efficient inference kernels.

Beyond GPTQ/AWQ, the PTQ landscape includes methods that target different bottlenecks:
SmoothQuant addresses activation outliers by migrating quantization difficulty from activations to weights via a mathematically equivalent rescaling transformation, enabling efficient W8A8 inference while preserving accuracy \citep{smoothquant}. OmniQuant optimizes quantization parameters (e.g., clipping thresholds and equivalent transformations) in a data-efficient manner to approach QAT-like performance while retaining PTQ efficiency \citep{omniquant2023}. Many additional variants exist (e.g., alternative rounding, clipping, mixed precision, and KV-cache/activation quantization), but a recurring theme is that PTQ aims to \emph{approximate} the full-precision model with minimal adaptation cost.

\section{Metrics}
\label{sec:metrics}

We evaluate quantized reasoning models along two axes: task performance and reasoning efficiency.
\begin{itemize}[leftmargin=*]
    \item \textbf{Task performance.} We follow each benchmark's official protocol. By default we report exact-match accuracy; for math benchmarks we normalize the output formatting before comparing the extracted final answer to the reference, and for code benchmarks we report Pass@1.
    \item \textbf{Reasoning and answer length.} Reasoning model outputs typically consist of a chain-of-thought (CoT) trace followed by an (often shorter) final answer. We report the token counts of the two portions separately, which lets us distinguish changes in reasoning length from changes in answer length.
\end{itemize}

\section{Prompt based methods}
\label{app:prompts}
We evaluate three prompt-based interventions that aim to suppress redundant reasoning, encourage token-efficient reasoning, and directly discourage explicit CoT generation. Here is the detail of the prompts.
\begin{itemize}
    \item \textit{do not think}: During reasoning, do not think.
    \item \textit{less tokens}: During reasoning, use as few tokens as possible.
    \item \textit{no restate}: During reasoning, do not restate previous steps unless necessary for correctness. Each step should introduce a new inference, calculation, or decision. If the current thought does not add new information, stop that line of reasoning and move on.
\end{itemize}

\section{Additional evaluation details}
\label{app:eval_details}

This appendix provides benchmark descriptions, prompt templates, decoding settings, and additional evaluation details.

\subsection{Benchmarks and evaluation protocol}
\label{app:benchmarks}

Table~\ref{tab:benchmark_details} summarizes the key characteristics of the benchmarks used in our evaluation.

\begin{table*}[t]
\centering
\scriptsize
\setlength{\tabcolsep}{6pt}
\renewcommand{\arraystretch}{1.15}
\resizebox{\textwidth}{!}{%
\begin{tabular}{l|l|l|c|c|l}
\toprule
\textbf{Benchmark} & \textbf{Version} & \textbf{Domain} & \textbf{\# Problems} & \textbf{Evaluation} & \textbf{Notes} \\
\midrule
AIME25 & 25 & Math Competition & 30 & Exact Match & American Invitational Mathematics Examination 2025 \\
GSM8K & --- & Math Reasoning & 1,319 & Exact Match & Grade-school math word problems \\
LiveCodeBench & V5, V6 & Code Generation & 182 & Pass@k & Live coding evaluation benchmark \\
MBPP & --- & Code Generation & 974 & Pass@k & Mostly Basic Python Programming benchmark \\
GPQA-Diamond & --- & Science Reasoning & 198 & Accuracy & Diamond subset of graduate-level Google-proof Q\&A benchmark \\
BBH & --- & General Reasoning & 6,511 & Exact Match / Accuracy & 23 challenging BIG-Bench Hard tasks \\
HellaSwag & --- & Commonsense Reasoning & 10,042 & Accuracy & Commonsense natural language inference benchmark \\
PIQA & --- & Physical Commonsense & 1,838 & Accuracy & Physical interaction question answering benchmark \\
MMLU-Pro & --- & Multitask Knowledge & 12,032 & Accuracy & More challenging multitask language understanding benchmark \\
\bottomrule
\end{tabular}%
}
\caption{Benchmark details for our evaluation. Benchmarks span mathematical reasoning, code generation, science reasoning, general reasoning, commonsense reasoning, and multitask knowledge evaluation.}
\label{tab:benchmark_details}
\end{table*}

We use the default prompt for reasoning models. More specifically: \\
``Please reason step by step, and put your final answer within \texttt{\textbackslash boxed\{\}}.''

\subsection{Model details}
\label{app:model}
Table~\ref{tab:model_details} lists the evaluated checkpoints, model metadata, and decoding settings used throughout the paper.

\begin{table*}[t!]
\centering
\scriptsize
\setlength{\tabcolsep}{5pt}
\renewcommand{\arraystretch}{1.15}
\resizebox{\textwidth}{!}{%
\begin{tabular}{l|l|l|c|c|c|c|c|c}
\toprule
\textbf{Model} & \textbf{Organization} & \textbf{Size} & \textbf{Arch.} & \textbf{Release} & \textbf{Max seq. len} & \textbf{Temp.} & \textbf{Top-$k$} & \textbf{Top-$p$} \\
\midrule
\texttt{Qwen3-4B-Thinking-2507} & Alibaba (Qwen Team) & 4B & Dense & 2025-07 & 256K & 0 & 20 & 0.95 \\
\texttt{Qwen3-30B-A3B-Thinking-2507} & Alibaba (Qwen Team) & 30B (A3B active) & MoE & 2025-07 & 256K & 0 & 20 & 0.95 \\
\texttt{Phi-4-Reasoning-Plus} & Microsoft & 14B & Dense & 2025-04 & 32K & 0 & 50 & 0.95 \\
\bottomrule
\end{tabular}%
}
\caption{Model metadata and decoding settings used in our evaluation. ``Max seq.\ len'' denotes the maximum supported context length; Since temperature is set to 0, top-k/top-p do not affect greedy decoding.}
\label{tab:model_details}
\end{table*}

\paragraph{Hardware.}
All experiments (PTQ and evaluation) are run on NVIDIA H100 GPUs.

\section{Correctness-Conditioned Token Inflation}
\begin{figure*}[t]
  \centering
  \includegraphics[width=\textwidth]{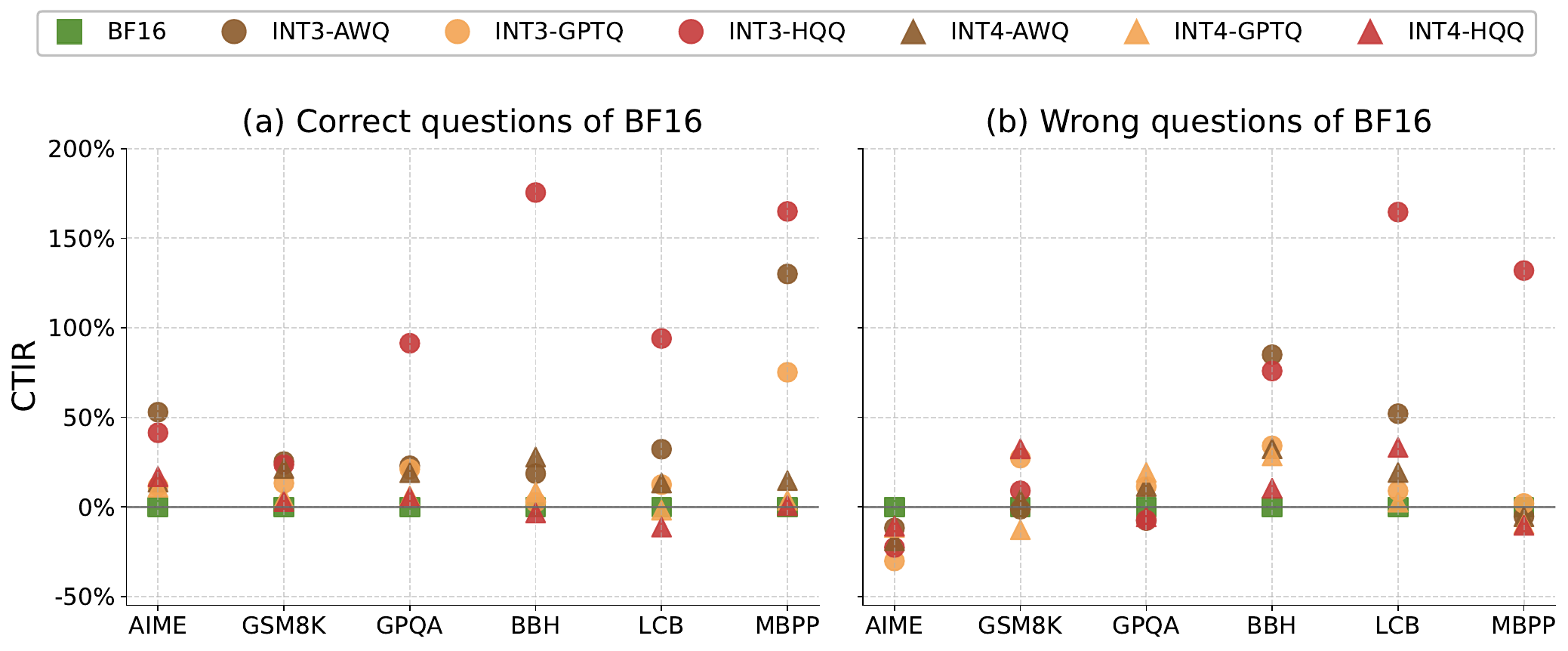}
  \vspace{-20pt}
  \caption{Correctness-conditioned CTIR for Qwen3-30B. We split questions by whether BF16 answers them correctly and compute token inflation within each subset.}
  \label{fig:correct_wrong}
  \vspace{-10pt}
\end{figure*}

\begin{figure*}[t]
  \centering
  \includegraphics[width=\textwidth]{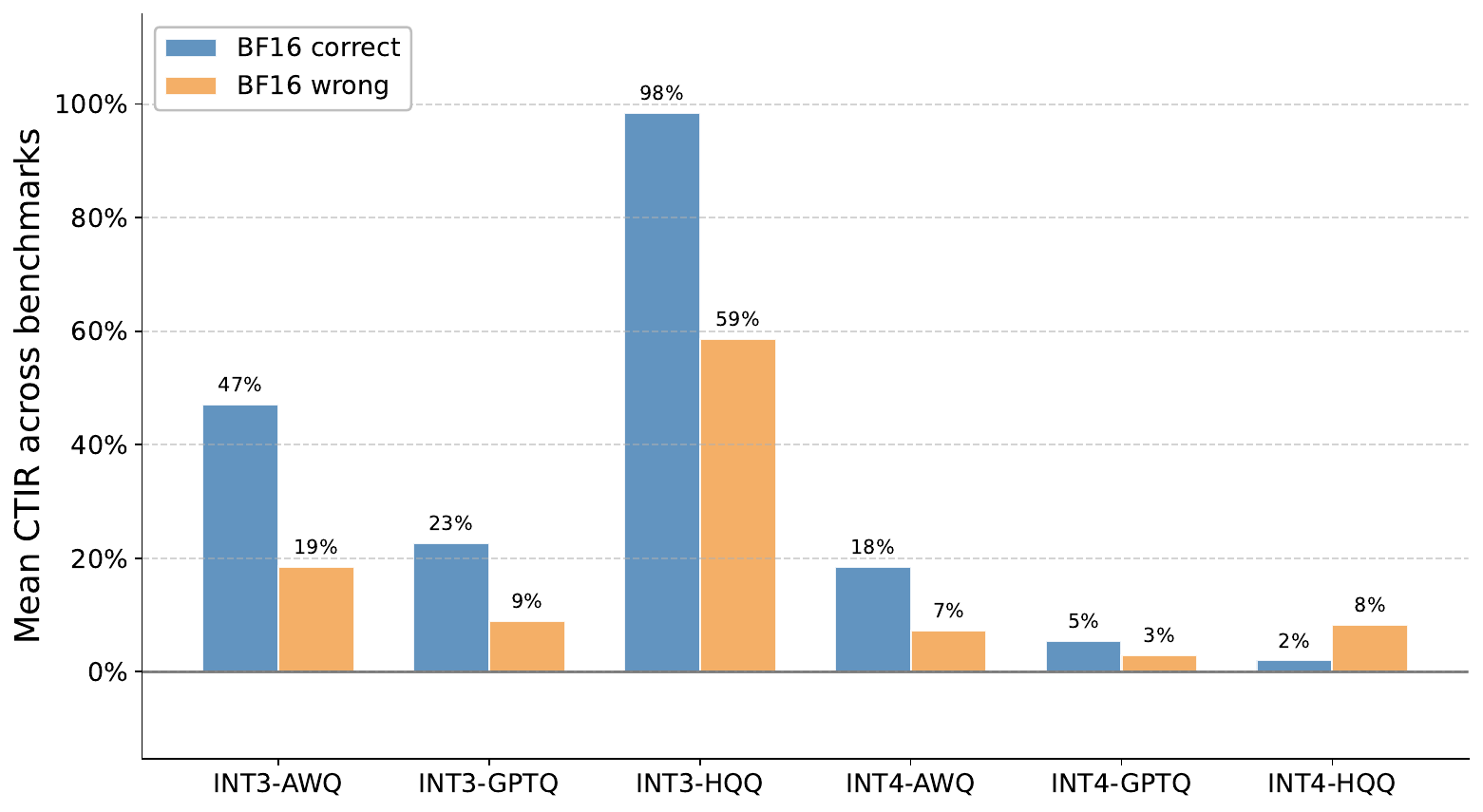}
  \vspace{-20pt}
  \caption{Correctness-conditioned CTIR averaged over benchmarks.}
  \label{fig:correct_wrong_bar}
  \vspace{-10pt}
\end{figure*}

To further analyze where quantization-induced token inflation occurs, we conduct a correctness-conditioned experiment on Qwen3-30B with reasoning tasks, by partitioning evaluation instances according to whether the BF16 reference model answers them correctly or incorrectly. For each partition, we compute token inflation by comparing the CoT length of the quantized model with that of the BF16 model on the same set of instances. As shown in \fref{fig:correct_wrong}, token inflation is stronger and more consistent on the BF16-correct subset. In contrast, the BF16-wrong subset exhibits weaker and more irregular length changes, and several quantization settings even produce token deflation. This indicates that token inflation is not merely caused by already-failed or unstable examples becoming longer under quantization. Instead, quantization tends to extend reasoning trajectories on instances where the full-precision model has a valid solution path, suggesting that low-bit quantization can make the model spend more intermediate reasoning tokens to preserve or recover the same reasoning outcome. We also present the comparison of different methods, as shown in \fref{fig:correct_wrong_bar}. The method-level results in \fref{fig:correct_wrong_bar} make the correctness--CTIR relationship more explicit. For nearly all quantization configurations, BF16-correct instances show higher CTIR than BF16-wrong instances, and the gap is especially large under more aggressive 3-bit quantization. Thus, quantization-induced token inflation appears to arise when the model has a valid reasoning path but needs additional intermediate computation under low-bit perturbations to preserve the final answer.

\end{document}